\documentclass[runningheads]{llncs}

\PassOptionsToPackage{pagebackref,breaklinks,colorlinks,citecolor=eccvblue}{hyperref}

\usepackage[year=2026,ID=8765]{eccv} 

\usepackage{eccvabbrv}
\usepackage{graphicx}
\usepackage{booktabs}
\usepackage[accsupp]{axessibility}
\usepackage{hyperref}
\usepackage{orcidlink}
\usepackage{amsmath}
\usepackage{amssymb}
\usepackage{placeins}
\usepackage{float}
\usepackage{wrapfig}
\usepackage{caption}
\usepackage{siunitx}  
\usepackage{cellspace} 
\usepackage[table]{xcolor}
\usepackage{tabularx}
\usepackage{multirow}

\definecolor{ourscolor}{HTML}{E6F2FF}

\begin{document}


\title{FlexComposer: Unified Video Compositing \\ from Images to Dynamic
Footage \\ with Flexible Trajectory Control}

\titlerunning{Flexcomposer}


\author{Songchun Zhang\inst{1} \and Sitong Guo\inst{2} \and Xianghao Kong\inst{1} \and Pengwei Liu\inst{2} \and Yuwei Guo\inst{3} \and Lvmin Zhang\inst{4} \and Anyi Rao\inst{1}}
\authorrunning{Songchun Zhang et al.}
\institute{
  $^1$HKUST, $^2$ZJU, $^3$CUHK, $^4$Stanford University \\
  \href{https://franklinz233.github.io/projects/flexcomposer/}{Project Page}
}

\maketitle

\begin{figure*}[h] 
  \centering
  \vspace{-0.8cm}
  \includegraphics[width=\textwidth]{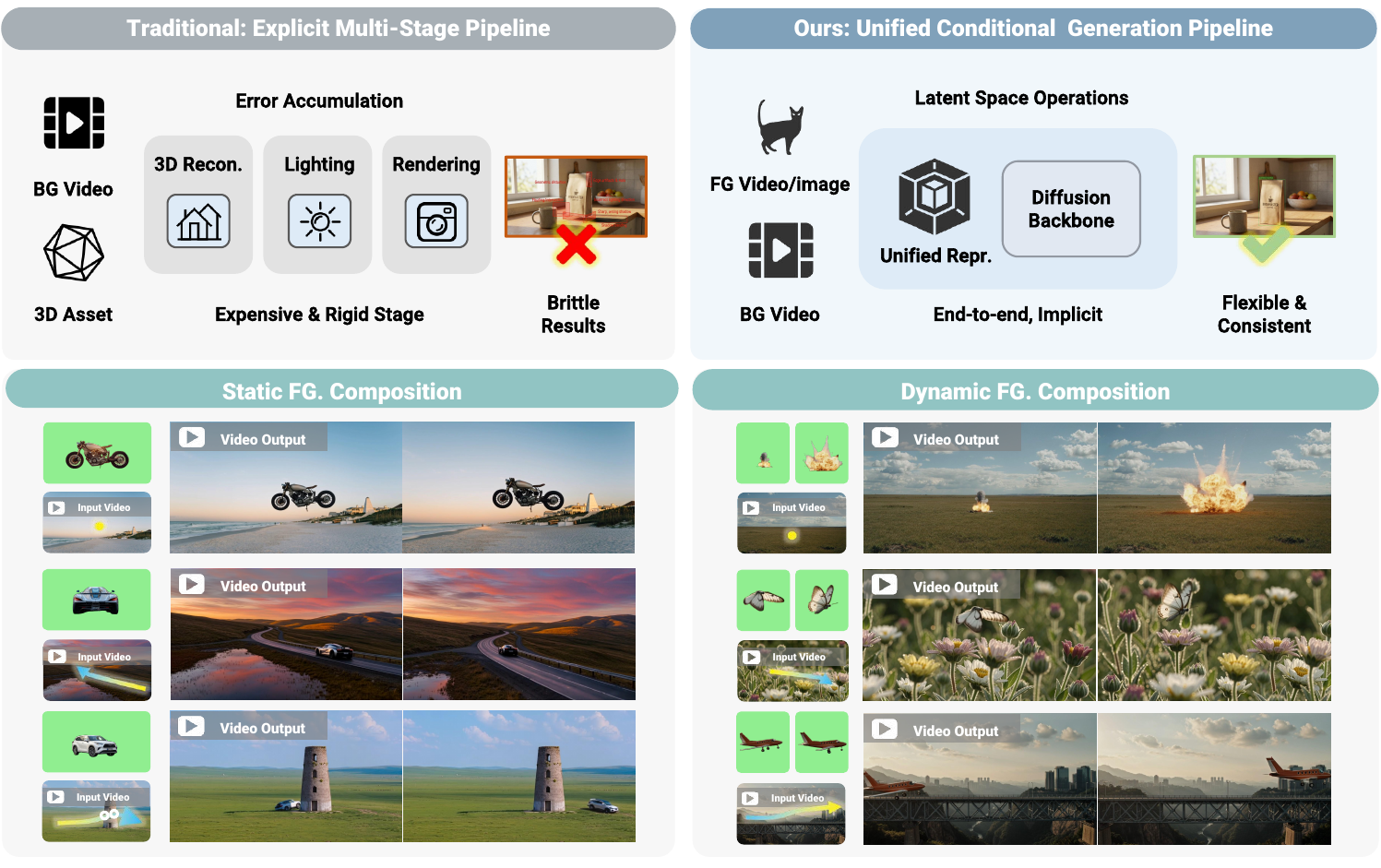}
  \caption{\textbf{Versatile Video Compositing with FlexComposer.} Our framework unifies the insertion of diverse assets into target videos through user-defined trajectories (visualized as arrows in input thumbnails).
  \textbf{Left (Static Foreground Comp.):} Given a single static image, the model synthesizes realistic object motion and view changes while achieving photorealistic lighting adaptation (middle row) and handling depth occlusions (bottom row).
  \textbf{Right (Dynamic Foreground Comp.):} For video inputs, FlexComposer preserves the internal dynamics of the asset (e.g., the expansion of the explosion or the flapping of butterfly wings) while harmonizing it spatially and photometrically with the background scene.}
  \label{fig:teaser}
\end{figure*}

\vspace{-2em}

\FloatBarrier
\begin{abstract}
Generative video compositing, which involves inserting external assets seamlessly into existing video sequences, is essential for content creation and visual effects. 
However, existing approaches suffer from a control-fidelity trade-off: they either hallucinate motion from static images, failing to preserve the dynamics of pre-animated assets, or lack fine-grained spatial control for precise asset placement along user-defined trajectories. 
We propose FlexComposer, a unified framework that standardizes video compositing as a trajectory-guided conditional generation task, enabling the seamless integration of both static images and dynamic footage.
Our approach introduces three key designs: (1) a Unified Canonical Foreground Representation that decouples an object's intrinsic motion from its global displacement, standardizing heterogeneous inputs into a stabilized, centered latent space; (2) a Spatial-Aware Latent Injection strategy that exploits the translation equivariance of VAE latent spaces to transport canonical features onto target trajectories via a parameter-free mechanism; and (3) a Hybrid Dataset and Synthetic-to-Real Curriculum that synergizes procedural simulation, real-world cinematic footage, and generative data to implicitly learn physically plausible illumination and shadow harmonization.
This unified design handles diverse inputs—from product photos to dynamic subjects—achieving high-fidelity motion control and environmental integration without the need for explicit 3D reconstruction or auxiliary learnable adapters. 
Extensive experiments demonstrate that FlexComposer outperforms state-of-the-art methods in visual quality, temporal consistency, and trajectory adherence.

\end{abstract}

\section{Introduction}
The realistic insertion of virtual assets into existing video sequences is a fundamental yet challenging problem in computer vision and graphics.
This capability serves as the backbone for visual effects, augmented reality, 3D-aware content creation~\cite{zhang20243d,zhang2025spatialcrafter,zhang2025pragmatist}, and the burgeoning field of personalized video content creation.
%
Achieving seamless composites requires simultaneously preserving background integrity, maintaining spatiotemporal consistency of inserted foregrounds, and enabling precise motion control over their trajectories.

While traditional graphics pipelines~\cite{karsch2014automatic} and harmonization techniques~\cite{vtt,ke2022harmonizer} excel in rendering or color adjustment, they operate via explicit decoupled stages prone to cascading error accumulation, inherently struggling to integrate dynamic assets with coherent motion. 
In contrast, recent generative frameworks~\cite{ju2025editverse,yang2025unified,ye2025unic,team2025klingo1,chen2026vino,cai2025omnivcus,tu2025videoanydoor,vace,kong2026composing} leverage diffusion models to address these limitations.
These approaches offer promising avenues for composition, ranging from zero-shot subject animators~\cite{tu2025videoanydoor,cai2025omnivcus} to unified in-context editors~\cite{ju2025editverse,wei2025univideo,chen2026vino,ye2025unic} and large-scale foundation models~\cite{team2025klingo1,yang2025unified}. 
However, despite their versatility, a critical gap remains: most methods either hallucinate motion from static references~\cite{tu2025videoanydoor} or operate at abstract semantic levels~\cite{ju2025editverse,wei2025univideo}. Crucially, they lack a unified mechanism to seamlessly integrate both static images and dynamic video assets—preserving intrinsic fidelity while ensuring fine-grained spatial control.

To address these limitations, we propose FlexComposer, a unified conditional video compositing framework. We bridge the gap between control and fidelity through three key technical components.
First, to handle heterogeneous inputs, we introduce a Unified Canonical Foreground Representation that decouples intrinsic motion from global displacement by stabilizing dynamic videos into canonical sequences at the pixel level, enabling consistent treatment of diverse inputs. 
Second, to achieve precise spatial control without auxiliary adapters (e.g., ControlNet~\cite{zhang2023adding}), we propose Spatial-Aware Latent Injection, a parameter-free mechanism that exploits VAE latent translation equivariance to directly transport canonical features onto user-defined trajectories while preserving temporal synchronization. 
Third, to resolve photometric inconsistencies, we construct a Hybrid Dataset combining procedural simulation data for geometric grounding, real-world footage for realism, and generative data for semantic diversity, trained through a synthetic-to-real curriculum. 
Extensive experiments demonstrate that FlexComposer achieves superior performance in visual quality, temporal consistency, and controllability.

In summary, our contributions are threefold: 
(1) we propose \textbf{FlexComposer}, a unified generative framework that seamlessly integrates both static images and dynamic video clips into target sequences by utilizing a canonical representation to decouple intrinsic object motion from global trajectory control; 
(2) we introduce \textbf{Spatial-Aware Latent Injection}, a parameter-free mechanism that leverages translation equivariance to enable precise spatial control while preserving the high-fidelity details of foreground assets without auxiliary adapters; and 
(3) we implement a \textbf{synthetic-to-real curriculum} learning strategy that leverages a diverse mixture of procedural simulation and real-world footage. This approach empowers the model to robustly generalize across diverse scenes with coherent synthesis of lighting and shadows.
Extensive experiments demonstrate that FlexComposer achieves superior performance in visual quality, temporal consistency, and controllability.
\section{Related Work}

\subsection{Video Generative Models}
The landscape of video generation has shifted fundamentally from early GANs~\cite{goodfellow2014generative} and autoregressive transformers~\cite{esser2021tamingvqgan} to Diffusion Models (VDMs)~\cite{ho2022video, blattmann2023stable, ho2020denoising}, which now serve as the gold standard for high-fidelity generation. 
While Latent Diffusion Models (LDMs)~\cite{rombach2022high} balanced quality and efficiency, the recent emergence of Video Diffusion Transformers (DiTs)~\cite{peebles2023scalablediffusionmodelstransformers,yang2024cogvideox,kong2024hunyuanvideo}, autoregressive video models~\cite{zhang2026astrolabe}, and memory-augmented long video generation~\cite{bian2026echo}, exemplified by Sora~\cite{brooks2024sora,opensora,keling}, has demonstrated superior scalability and temporal coherence.
Despite their ability to generate photorealistic content from text, these foundation models inherently lack precise spatial controllability. 
Our work builds upon the powerful priors of these large-scale DiTs, bridging the gap between high-fidelity synthesis and fine-grained, user-specified control.

\subsection{Motion-Controlled Video Generation}
While T2V models achieve photorealism, text prompts lack the granularity required for precise control. 
Existing spatial guidance methods fall into three categories.
First, trajectory-based control employs sparse 2D drag points or bounding boxes~\cite{yin2023dragnuwa,wang2024motionctrl,huang2023fine,guo2024sparsectrl,namekata2024sg,yu2024zero,mou2024revideo,pandey2024motion}, with recent works integrating trajectory adapters directly into DiT architectures~\cite{geng2024motion,zhang2025tora,chu2025wanmove,wang2025world,wang2024boximator,ma2024trailblazer}. However, these methods operate in 2D screen space and fail to comprehend the underlying 3D scene geometry.
To incorporate 3D geometric awareness, efforts have been directed toward camera control and 3D-consistent scene generation. Approaches range from traditional reconstruction-based methods~\cite{gao2021dynamic,lee2024fast,xian2021space,wang2025shape} to generative techniques using multi-view diffusion, inpainting, pose conditioning, or sparse-view reconstruction~\cite{zhang20243d,zhang2025spatialcrafter,zhang2025pragmatist,bahmani2025lyra,bai2024syncammaster,sun2025dimensionx,wang20254real,wu2025cat4d,zhou2025stable,ma2025follow,ren2025gen3c,you2024nvs,yu2025trajectorycrafter,yu2024viewcrafter,bai2025recammaster,he2024cameractrl}.
Nevertheless, such global control focuses on the ego-motion rather than the fine-grained spatial dynamics of individual objects within the scene.
Alternatively, structure-based control leverages dense priors such as skeleton, depth, or normal maps~\cite{ma2024follow,lei2024animateanything,wang2024videocomposer,gu2025diffusiondas}. While offering geometric guidance, these methods impose heavy data preparation burdens that limit their applicability.

\subsection{Video Compositing and Harmonization}
Traditional approaches to video composition employ explicit decoupled pipelines. Graphics-based methods~\cite{karsch2014automatic} rely on 3D reconstruction, lighting estimation, and ray tracing in separate stages, with error accumulation across stages limiting robustness. 
Professional tools~\cite{foundry_nuke, adobe_after_effects} built on Porter and Duff's compositing operator~\cite{porter1984compositing} enable nondestructive layer-based workflows but require extensive manual per-frame artistry for masking and effect tuning.
Recent lightweight approaches focus on color harmonization~\cite{vtt, yang2025gencompositor}, enabling rapid integration through color matching and tone adjustment.
However, these methods operate in a rigid copy-paste paradigm without accounting for geometric interactions, lighting adaptation, or dynamic foreground motion—prerequisites for realistic asset composition.
Instruction-based video editing originated from image-level methods~\cite{brooks2023instructpix2pix} and has evolved toward video through per-video fine-tuning~\cite{wu2023tune,qi2023fatezero}, inference-only protocols~\cite{wu2025insvie,ku2024anyv2v,cheng2023consistentinsv2v,mai2025easyv2v,ju2025editverse,yang2025unified,ye2025unic,team2025klingo1,chen2026vino,cai2025omnivcus}, and concept-level image/video composition~\cite{kong2026composing}.
While diffusion priors enable semantic editing, they lack precise trajectory mechanisms for asset insertion. 
Conversely, generative compositing frameworks~\cite{tu2025videoanydoor,vace,cai2025omnivcus} address insertion but primarily hallucinate motion from static references, failing to preserve the intrinsic dynamics of pre-animated assets. 
Our work bridges these complementary research directions by unifying foreground representation with spatial control mechanisms within a generative framework specifically designed for composition.
\section{Method}
\begin{figure*}[t!]
    \centering
    \includegraphics[width=1.0\linewidth]{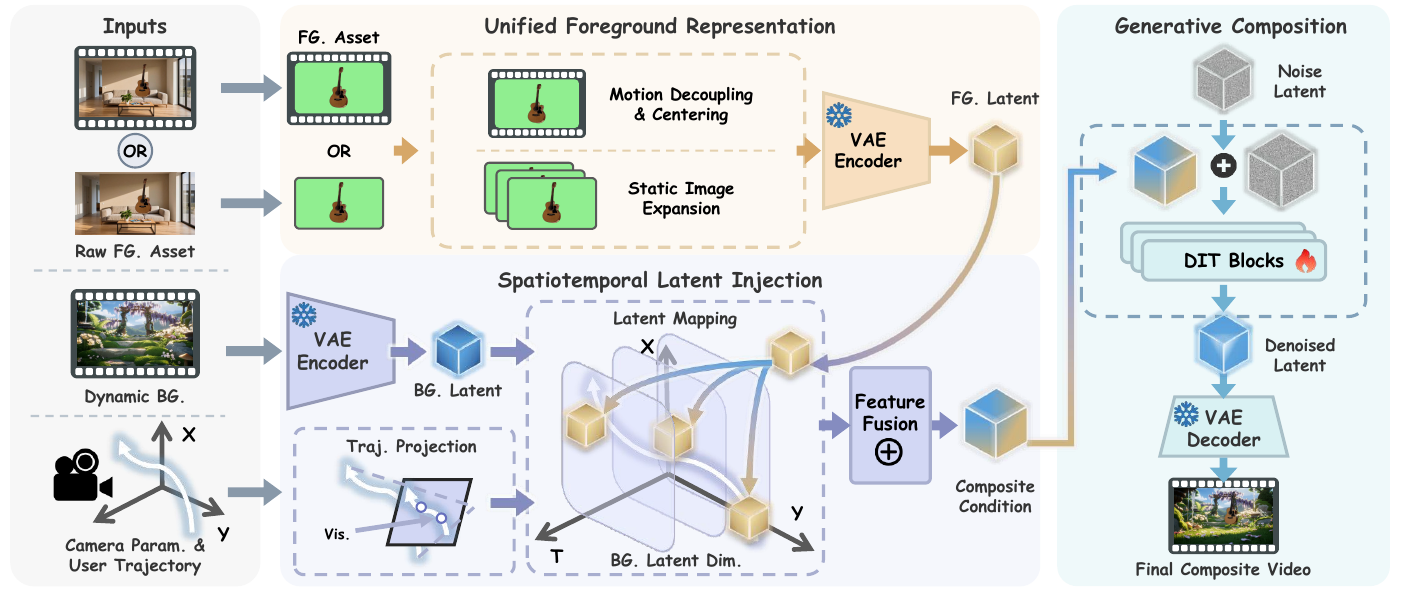}
    \vspace{-2mm}
    \caption{\textbf{The Pipeline of FlexComposer.} (1) \textbf{Inputs}: A dynamic background, a static/dynamic foreground asset, a 3D trajectory, and visibility masks. (2) \textbf{Unified Representation}: Diverse inputs are encoded into a canonical feature space to preserve identity and local motion. (3) \textbf{Spatiotemporal Latent Injection}: Canonical features are warped into the background latent space along the trajectory, utilizing a visibility gate for occlusions. (4) \textbf{Generative Composition}: The warped features condition a video diffusion model to synthesize the final composite.}
\vspace{-2mm}
\label{fig:pipeline}
\end{figure*}

Given a foreground asset $V_{raw}$ and a background sequence $V_{bg}$, our goal is to synthesize a composite video where the asset follows a user-defined trajectory $\mathcal{T}_{user}$. 
Users can optionally specify a visibility schedule $\mathcal{V}_{user}[t] \in \{0, 1\}$ to control when the object appears or disappears.
As illustrated in Figure.~\ref{fig:pipeline}, our framework follows a three-stage pipeline: 
(1) Unified Canonical Foreground Representation, which standardizes diverse inputs into a stabilized, centered latent space; 
(2) Spatial-Aware Latent Injection, which transports canonical features onto the target trajectory via a parameter-free mechanism; and (3) Generative Composition, which leverages a conditional diffusion transformer to synthesize the final video.

\subsection{Unified Canonical Foreground Representation}
We unify heterogeneous inputs ($V_{\text{raw}}$) into a canonical representation through pixel-space preprocessing followed by latent encoding.

\noindent\textbf{Dynamic Video Stabilization.}
To eliminate conflicting camera ego-motion, we utilize Grounded-SAM2~\cite{kirillov2023segment} for segmentation and SpatialTracker v2~\cite{Xiao2025SpatialTrackerV23P} for centroid tracking ($\mathbf{c}_t$) and visibility estimation ($\mathcal{V}_{\text{track}}$). 
We apply a reverse translation to align $\mathbf{c}_t$ to the canvas center, yielding a stabilized, in-place sequence $\mathbf{I}_{\text{stab}}$ where the subject animates at the center, effectively decoupling local motion from global displacement.

\noindent\textbf{Static Image Expansion.}
For static inputs, we replicate the image to target duration $T$ to form $\mathbf{I}_{\text{expand}}$. We further inject Gaussian noise along the temporal dimension after encoding, encouraging the generative backbone to hallucinate plausible temporal dynamics while preserving the subject's identity.

\noindent\textbf{Relighting Augmentation.}
To prevent the network from merely copy-pasting the source appearance and to enforce photometric consistency with diverse background contexts, we apply random illumination perturbations to the frames via a relighting module~\cite{liu2025unilumos}. 
The final preprocessed sequence is denoted as $\mathbf{I}_{\text{can}}$.

\noindent\textbf{Latent Space Encoding.}
We encode the preprocessed sequence $\mathbf{I}_{\text{can}}$ using the Video VAE encoder: $\mathbf{Z}_{\text{can}} = \mathcal{E}(\mathbf{I}_{\text{can}}) \in \mathbb{R}^{T' \times H' \times W' \times C}$, where $T' = T / f_t$, $H' = H / f_s$, $W' = W / f_s$ reflect temporal and spatial downsampling factors. 
For static inputs, we inject Gaussian noise along the temporal axis to break frame-wise
redundancy and encourage motion synthesis

\subsection{Spatial-Aware Latent Injection}
\label{sec:method_injection}

Unlike existing methods~\cite{gu2025diffusiondas,geng2024motion} that rely on auxiliary adapters which often suffer from signal degradation, we introduce a parameter-free injection strategy.

\noindent\textbf{Spatiotemporal Coordinate Mapping.}
We first project the 3D trajectory $\mathcal{T}_{3D} = \{\mathbf{P}_k\}_{k=1}^{T}$ onto the 2D image plane to obtain pixel coordinates $\mathbf{p}_k = \Pi(\mathbf{P}_k)$, where $\Pi(\cdot)$ incorporates the camera intrinsics and extrinsics.
To align these coordinates with the latent space of VAE, which undergoes spatial downsampling by factor $f_s$ and temporal compression by factor $f_t$, we compute the discrete latent coordinate $\tilde{\mathbf{u}}_n$ for each latent frame index $n$:
\begin{equation}
\tilde{\mathbf{u}}_n = \left\lfloor \frac{1}{f_t} \sum_{k=(n-1)f_t + 1}^{n \cdot f_t} \frac{\mathbf{p}_k}{f_s} \right\rceil, \quad 1 \leq n \leq \frac{T}{f_t}
\label{eq:latent_coord}
\end{equation}
where $\lfloor \cdot \rceil$ denotes the nearest integer rounding operation. 
This discretization ensures that the continuous projected coordinates are accurately mapped to the fixed integer grid indices of the latent feature map $\mathbf{Z}$.

\noindent\textbf{Frame-wise Feature Transport.}
We define a transport operation $\Psi$ that maps features from the canonical center $\mathbf{c}$ to the target trajectory $\tilde{\mathbf{u}}_n$.
Let $\Omega$ denote the spatial extent (bounding box) of the foreground object in the latent space.
For every local spatial offset $\boldsymbol{\delta} \in \Omega$, we map the feature at the canonical position $\mathbf{c} + \boldsymbol{\delta}$ to the target position $\tilde{\mathbf{u}}_n + \boldsymbol{\delta}$ in the composite latent $\mathbf{Z}_{\text{trans}}$:
\begin{equation}
\mathbf{Z}_{\text{trans}}[n, \tilde{\mathbf{u}}_n + \boldsymbol{\delta}] = \mathbf{Z}_{\text{can}}[n, \mathbf{c} + \boldsymbol{\delta}] \cdot \mathcal{V}[n, \boldsymbol{\delta}]
\label{eq:transport}
\end{equation}
where $\mathcal{V}[n, \boldsymbol{\delta}] = \mathcal{V}_{\text{vis}}[n] \cdot \alpha[n, \boldsymbol{\delta}] \in \{0, 1\}$ combines a temporal visibility flag with the spatial segmentation mask $\alpha[n, \boldsymbol{\delta}]$.
During training, $\mathcal{V}_{\text{vis}}$ is derived from SpatialTracker's occlusion detection; at inference, it is user-specified via $\mathcal{T}_{user}$, enabling explicit control over when the object appears along the trajectory.
Crucially, by coupling the temporal index $n$ on both sides of Eq.~\ref{eq:transport}, we preserve the object's intrinsic dynamics while updating its global location.

\subsection{Generative Composition}
\label{sec:method_generation}

Our video synthesis leverages the Wan Video Diffusion Transformer~\cite{wan2025wan} as the generative backbone. 
However, the standard I2V protocol is insufficient for our dynamic compositing task that requires dense spatiotemporal guidance. 
We therefore adapt the conditioning mechanism to accept our composite features.

\noindent\textbf{Composite Feature Fusion.}
We fuse the transported foreground features $\mathbf{Z}_{\text{trans}}$ with the background latent $\mathbf{Z}_{\text{bg}}$ to construct the composite condition $\mathbf{C}_{\text{fuse}}$:
\begin{equation}
\mathbf{C}_{\text{fuse}} = \mathbf{Z}_{\text{bg}} \odot (1 - \mathbf{M}) + \mathbf{Z}_{\text{trans}} \odot \mathbf{M}
\end{equation}
where $\mathbf{M}$ denotes the foreground mask in latent space and $\odot$ represents element-wise multiplication.

\noindent\textbf{Conditional Input Formation.}
%
The network input is formed by channel-wise concatenating the noisy latent $\mathbf{x}_t$ at diffusion timestep $t$ with the composite condition: $\mathbf{Z}_{\text{in}} = \text{Concat}(\mathbf{x}_t, \mathbf{C}_{\text{fuse}})$.
This dense spatiotemporal conditioning explicitly guides generation with strong geometric priors from the transported features, unlike standard I2V methods that hallucinate motion from a single frame.

\noindent\textbf{Training Objective.}
We optimize the model using the Flow Matching objective~\cite{lipman2022flow}. 
Crucially, the velocity prediction network $\mathbf{v}_{\theta}$ now takes the concatenated latent $\mathbf{Z}_{\text{in}}$ as its spatial input:
\begin{equation}
\mathcal{L}(\theta) = \mathbb{E}_{t, \mathbf{x}_t, \mathbf{c}_{\text{txt}}} \left[ \left\| \mathbf{v}_{\theta}(\mathbf{Z}_{\text{in}}, t, \mathbf{c}_{\text{txt}}) - \mathbf{v}_t(\mathbf{x}_t) \right\|^2 \right]
\label{eq:loss}
\end{equation}
where $\mathbf{x}_t$ denotes the flow state at timestep $t$, $\mathbf{v}_t$ is the ground-truth velocity field, and $\mathbf{c}_{\text{txt}}$ represents the text embeddings injected via cross-attention. 
This formulation ensures the diffusion process is strictly conditioned on our trajectory-aligned composite features.

\subsection{Hybrid Dataset and Curriculum Learning}
\label{sec:method_dataset}

To achieve precise motion control and robust generalization, we introduce a hybrid dataset construction pipeline and a three-stage curriculum learning strategy. 
This design is crucial for progressively disentangling geometric motion from photometric appearance.
Additional details are provided in the supplementary materials.

\noindent\textbf{Simulation Bootstrap.}
We first utilize procedurally generated data from dynamic 3D characters~\cite{mixamo} and synthetic environments~\cite{greff2022kubric} to establish geometric grounding. 
We train the model for 2k steps on 12k synthetic clips with explicit 3D vertex trajectories. 
This strict supervision warms up the spatial injection mechanism, enforcing the network to adhere to rigid spatial transport priors before tackling appearance harmonization.

\noindent\textbf{Real-World Adaptation.}
To bridge the domain gap, we curate a hybrid dataset comprising 54k high-quality real-world clips, augmented with estimated 3D trajectories. 
In this phase, we fine-tune the model for 5k steps. Training on these canonicalized real sequences enables the model to disentangle local motion from global movement and synthesize coherent environmental interactions (e.g., shadows and lighting) that are absent in synthetic data.

\noindent\textbf{Open-Domain Refinement.}
Finally, to support open-domain content creation, we introduce a generative synthesis pipeline prompting foundation T2I/I2V models. 
We refine the model for 1k steps on 5k highly diverse clips with pseudo-ground-truth trajectories. This strategy effectively prevents overfitting to limited real-world categories and broadens semantic coverage for novel assets.

\section{Experiments}
\label{sec:experiments}

To evaluate our method, we consider its nature as both a trajectory control framework and a video compositing tool. 
Since standard benchmarks typically focus on isolated tasks, we adopt a structured evaluation protocol across three domains: (1) \textit{Static Foreground Compositing} against motion-controllable I2V methods; (2) \textit{Dynamic Foreground Compositing} against video editing baselines; and (3) \textit{Video Harmonization} against specialized harmonization algorithms.


\subsection{Implementation Details}
We implement FlexComposer building upon the Wan2.1-I2V-14B architecture~\cite{wan2025wan}, initializing the backbone with Wan-Move weights~\cite{chu2025wanmove} to leverage robust motion priors. 
Unlike methods relying on heavy auxiliary networks, our spatial injection uses a parameter-free mechanism to preserve feature fidelity. 
To adapt the backbone without catastrophic forgetting, we employ Low-Rank Adaptation (LoRA)~\cite{hu2022lora} ($r=64$) on the projection layers of all DiT blocks. 
Training is conducted on 32 NVIDIA A100 GPUs using FSDP and Ulysses sequence parallelism (degree 4), optimized via AdamW ($lr=1 \times 10^{-5}$) with a total batch size of 32. More details can be found in supplemental materials.

\begin{figure*}[t]
    \centering
    \includegraphics[width=1.0\linewidth]{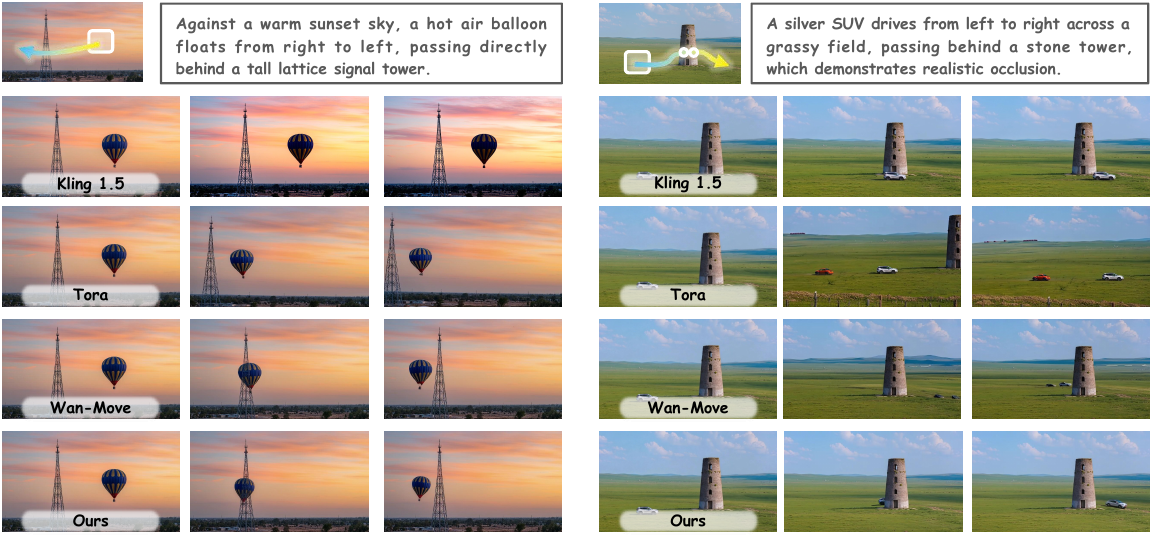}
    \caption{\textbf{Qualitative Comparison on Static Foreground Compositing.} We compare our method against trajectory-controlled I2V methods on two challenging scenarios. Our method demonstrates superior occlusion handling and trajectory adherence while maintaining visual fidelity}
    \vspace{-4mm}
    
\label{fig:exp_static}
\end{figure*}
\begin{figure*}[t!]
    \centering
    \includegraphics[width=1.0\linewidth]{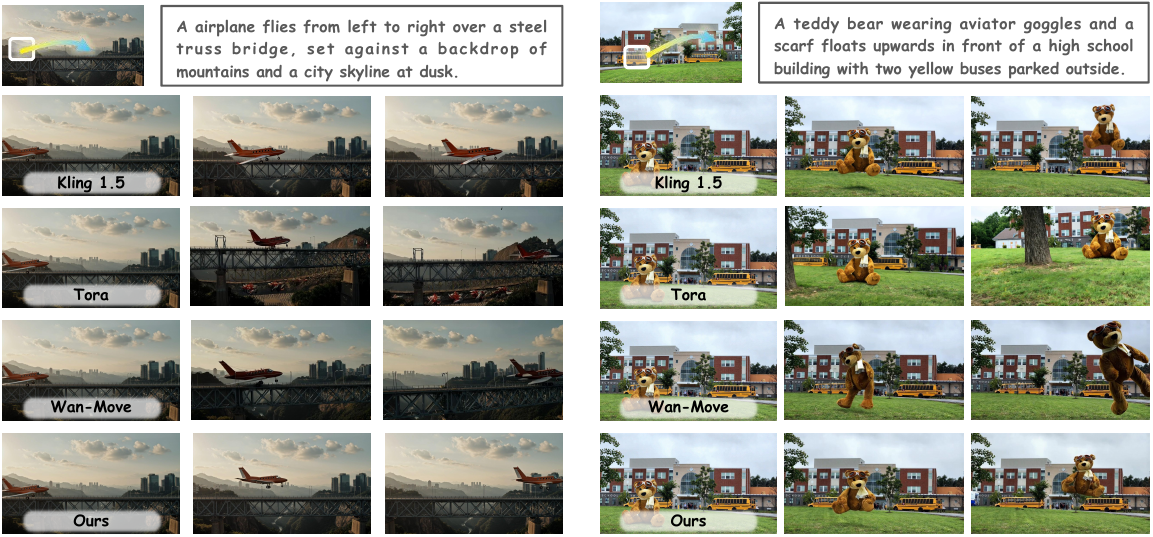}
    \caption{Additional Qualitative Comparison on Static Foreground Compositing. We compare our method against trajectory-controlled I2V methods on two challenging
scenarios.}
\label{fig:supp_3d_rot}
\end{figure*}

\begin{table}[t]
\centering
\caption{\textbf{Quantitative comparison on DAVIS.} Comparison of video quality (FID, FVD) and trajectory consistency (EPE) against recent baselines.}
\label{tab:static_comparision}
\begin{tabular}{l|ccccc}
\toprule
\textbf{Method} & \textbf{EPE} $\downarrow$ & \textbf{FID} $\downarrow$ & \textbf{FVD} $\downarrow$ & \textbf{PSNR} $\uparrow$ & \textbf{SSIM} $\uparrow$ \\
\midrule
ImageConductor~\cite{li2024image} & 14.92 & 54.5 & 512.8 & 11.55 & 0.468 \\
LeviTor~\cite{wang2025levitor}        & 3.65  & 22.3 & 116.1 & 13.20 & 0.515 \\
Tora~\cite{zhang2025tora}           & 3.52  & 26.1 & 128.5 & 13.65 & 0.492 \\
MagicMotion~\cite{li2025magicmotion}    & 3.48  & 24.5 & 112.9 & 12.90 & 0.535 \\
Wan-Move~\cite{chu2025wanmove}       & \underline{2.50}  & \underline{14.7} & 94.3  & 16.50 & 0.610 \\
\midrule
\textbf{Ours (I2V)} & 2.38 & 14.1 & \underline{89.5} & \underline{16.85} & \underline{0.625} \\
\textbf{Ours (V2V)} & \textbf{2.15} & \textbf{13.4} & \textbf{82.6} & \textbf{17.40} & \textbf{0.658} \\
\bottomrule
\end{tabular}
\end{table}

\begin{figure*}[t]
    \centering
    \includegraphics[width=1.0\linewidth]{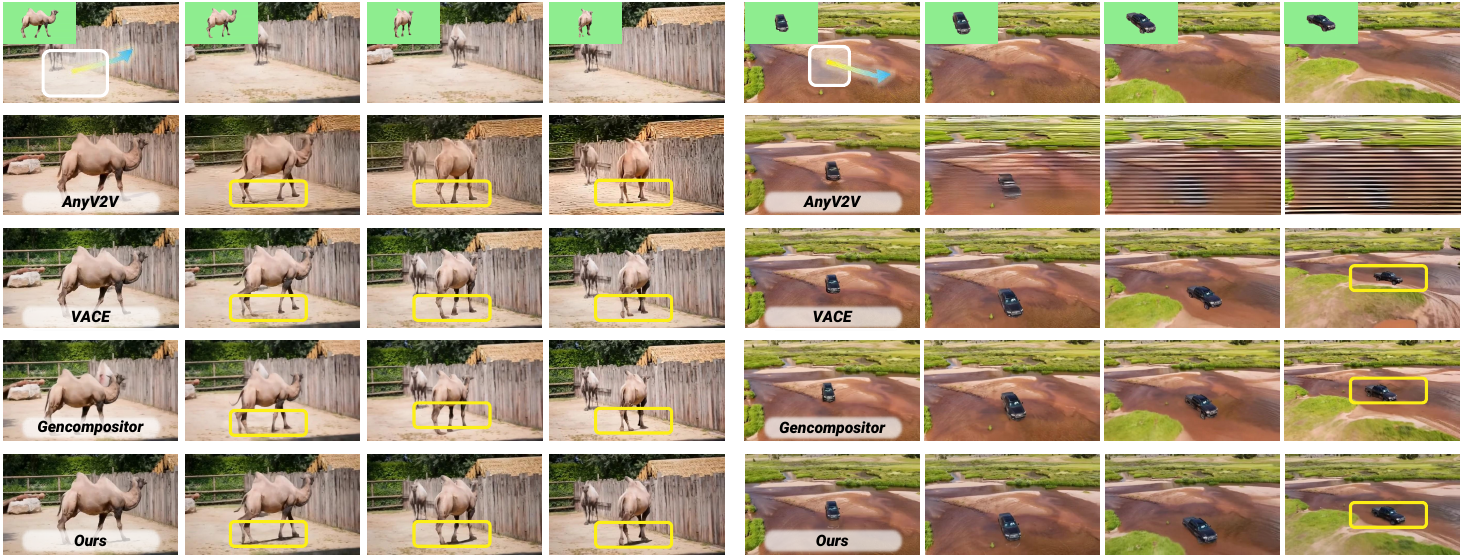}
    \vspace{-2mm}
    \caption{\textbf{Qualitative Comparison on Dynamic Foreground Compositing.} We evaluate our method against video editing and compositing approaches on the V2V task.}
    \vspace{-2mm}
\label{fig:dynamic}
\end{figure*}

\begin{table}[t]
\centering
\caption{\textbf{Quantitative comparison on the MoveBench dataset.} We report trajectory consistency and video quality metrics. Bold indicates the best results, and underline denotes the second best.}
\label{tab:movebench_results}
\begin{tabular}{l|ccccc}
\toprule
\textbf{Method} & \textbf{EPE} $\downarrow$ & \textbf{FID} $\downarrow$ & \textbf{FVD} $\downarrow$ & \textbf{PSNR} $\uparrow$ & \textbf{SSIM} $\uparrow$ \\
\midrule
ImageConductor~\cite{li2024image} & 15.55 & 34.8 & 422.5 & 13.50 & 0.485 \\
LeviTor~\cite{wang2025levitor}        & 3.45  & 18.3 & 99.2  & 15.55 & 0.538 \\
MagicMotion~\cite{li2025magicmotion}    & 3.25  & 17.6 & 97.1  & 14.85 & 0.562 \\
Tora~\cite{zhang2025tora}           & 3.32  & 22.8 & 101.5 & 15.65 & 0.548 \\
Wan-Move~\cite{chu2025wanmove}       & 2.60  & 12.2 & 83.5  & \underline{17.80} & \underline{0.640} \\
\midrule
\textbf{Ours (I2V)} & \underline{2.42} & \underline{11.9} & \underline{80.4} & 17.95 & 0.655 \\
\textbf{Ours (V2V)} & \textbf{2.38} & \textbf{11.2} & \textbf{74.9} & \textbf{18.60} & \textbf{0.682} \\
\bottomrule
\end{tabular}
\end{table}

\subsection{Static Foreground Compositing}
\label{sec:exp_static}

In this setting, we aim to animate a static image along a user-defined trajectory while maintaining identity.


\noindent\textbf{Setup.} 
We evaluate on DAVIS~\cite{wang2019learning} and MoveBench~\cite{chu2025wanmove}.
Following standard protocols, we evaluate the fidelity of generated videos using FVD~\cite{fvd} and FID~\cite{fid} for perceptual quality, alongside PSNR and SSIM~\cite{ssim} for frame-level accuracy.
We compare against state-of-the-art motion-controllable methods~\cite{chu2025wanmove,zhang2025tora,li2024image,li2025magicmotion,wang2025levitor} under two settings: \textit{I2V} (first frame only) and \textit{V2V} (full background context).



\noindent{{\textbf{Results and Analysis.}}}
Table~\ref{tab:static_comparision} summarizes the quantitative evaluation. 
In the I2V setting, our method demonstrates reliable trajectory adherence, comparing favorably to representative baselines such as ImageConductor~\cite{li2024image} and Tora~\cite{zhang2025tora}. 
These results suggest that our trajectory injection mechanism effectively guides motion generation, offering a robust alternative to pixel-level injection approaches for handling complex paths. 
Regarding visual quality, our method exhibits improved FVD and consistency. 
This indicates that full video context ensures temporal consistency, while our I2V variant maintains high fidelity without temporal priors.

\subsection{Dynamic Foreground Compositing}
\label{sec:exp_dynamic}

This task involves transforming a dynamic video foreground. 
The challenge lies in altering the global trajectory while preserving the object's intrinsic dynamics.

\noindent{{\textbf{Setup.}}}
We evaluate on a curated test set containing 50 video assets with diverse intrinsic motions and pre-defined trajectories.
We compare against Video Editing \& Compositing methods: AnyV2V~\cite{ku2024anyv2v}, VACE~\cite{vace}, and GenCompositor~\cite{yang2025gencompositor}.
We adopt the VBench~\cite{huang2024vbench} evaluation protocol, reporting Subject Consistency, Background Consistency, Motion Smooth, and FVD to comprehensively assess the quality of the composited videos.

\noindent{{\textbf{Results and Analysis.}}}
Table~\ref{tab:combined_comparison} displays the results. 
Video editing methods, which often prioritize texture editing, may sometimes distort motion patterns or fail to blend the subject seamlessly.
FlexComposer achieves the highest scores in Subject Consistency and Background Consistency, outperforming GenCompositor.
As shown in Figure~\ref{fig:dynamic}, we demonstrate compositing results on two scenarios.
Baseline methods exhibit various artifacts: AnyV2V occasionally loses subject coherence or introduces background inconsistencies, VACE struggles with temporal stability across frames, and GenCompositor sometimes alters the object's natural gait pattern or produces misaligned trajectories (see yellow bounding boxes in right columns).
In contrast, our method maintains both the intrinsic motion dynamics and precise trajectory adherence throughout the sequence, while seamlessly harmonizing the subject with the target background.


\begin{figure}[t]
    \centering
    \includegraphics[width=1.0\linewidth]{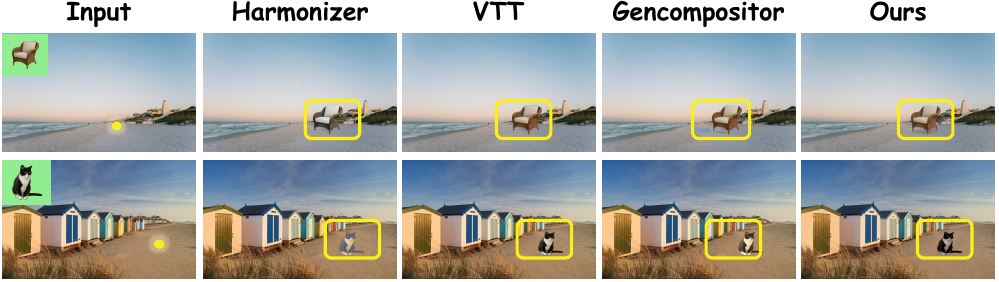}
    \caption{\textbf{Harmonization Results.} Our method synthesizes shadows and reflections alongside color adaptation.}
    \label{fig:harmonization_qualitative}
\end{figure}


\begin{table*}[t]
\centering
\caption{\textbf{Quantitative comparisons.}
\textit{Left:} V2V task using VBench metrics against recent video compositing and editing approaches.
\textit{Right:} Video Harmonization comparison of lighting and shadow adaptation quality.}
\label{tab:combined_comparison}

\begin{minipage}[t]{0.54\linewidth}
    \centering
    \resizebox{\linewidth}{!}{%
    \begin{tabular}{l|cccc}
        \toprule
        \textbf{Method} & \textbf{Sub. Cons.}$\uparrow$ & \textbf{BG. Cons.}$\uparrow$ & \textbf{Motion}$\uparrow$ & \textbf{FVD}$\downarrow$ \\
        \midrule
        AnyV2V~\cite{ku2024anyv2v}                & 88.02\% & 92.90\% & 96.85\% & 983.56 \\
        VACE~\cite{vace}                           & 89.51\% & 92.63\% & 98.21\% & 942.52 \\
        GenCompositor~\cite{yang2025gencompositor} & 89.75\% & 93.43\% & 98.69\% & 535.71 \\
        \midrule
        \textbf{Ours} & \textbf{91.20\%} & \textbf{93.85\%} & \textbf{98.71\%} & \textbf{468.30} \\
        \bottomrule
    \end{tabular}}
\end{minipage}
\hfill
\begin{minipage}[t]{0.42\linewidth}
    \centering
    \resizebox{\linewidth}{!}{%
    \begin{tabular}{lcccc}
        \toprule
        \textbf{Method} & \textbf{PSNR}$\uparrow$ & \textbf{SSIM}$\uparrow$ & \textbf{LPIPS}$\downarrow$ & \textbf{CLIP}$\uparrow$ \\
        \midrule
        Harmonizer~\cite{ke2022harmonizer}         & 39.8          & 0.942          & 0.041          & 0.961 \\
        VTT~\cite{vtt}                             & 40.1          & 0.931          & 0.046          & 0.956 \\
        GenCompositor~\cite{yang2025gencompositor} & \underline{42.0} & \underline{0.949} & \underline{0.039} & \underline{0.971} \\
        \midrule
        \textbf{Ours} & \textbf{42.5} & \textbf{0.952} & \textbf{0.037} & \textbf{0.974} \\
        \bottomrule
    \end{tabular}}
\end{minipage}

\end{table*}

\subsection{Video Harmonization}
\label{sec:exp_harmonization}

This task evaluates the adaptability of foreground lighting and shadows to the new background environment.

\noindent{{\textbf{Setup.}}}
We use the HYouTube~\cite{HYouTube} dataset, which contains paired unharmonized and harmonized examples.
We compare against Harmonizer~\cite{ke2022harmonizer}, VTT~\cite{vtt} and Gencompositor~\cite{yang2025gencompositor}, which are specialized for harmonization.
We report PSNR, SSIM~\cite{ssim}, LPIPS, and CLIP Score~\cite{hessel2022clipscorereferencefreeevaluationmetric} to evaluate the fidelity.

\noindent{{\textbf{Results and Analysis.}}}
As shown in Table~\ref{tab:combined_comparison}, FlexComposer demonstrates competitive performance across different metrics.
Figure~\ref{fig:harmonization_qualitative} presents two representative lighting scenarios: a coastal scene with warm sunset illumination (top) and a beach environment with directional lighting (bottom).
Specialized harmonization methods effectively adjust color tones to match ambient lighting conditions, showing their strength in appearance adaptation.
Our method attempts to complement color harmonization with geometric cues such as cast shadows.
While each method has its merits in different aspects of harmonization, our approach aims to balance both photometric and geometric consistency for composite video generation.

\begin{table}[t]
    \centering
  \caption{\textbf{Ablation Study.} Quantitative analysis of key components. We evaluate across different ablation variants.}
  \label{tab:ablation}
  \begin{tabular}{lccccc}
    \toprule
    \textbf{Variant} & \textbf{EPE} $\downarrow$ & \textbf{FVD} $\downarrow$ & \textbf{Motion} $\uparrow$ & LPIPS $\downarrow$ & \textbf{CLIP} $\uparrow$ \\
    \midrule
    w/o Canonical Repr. & 4.87 & 145.3 & 95.1 & 0.068 & 0.945 \\
    w/o Static Expansion & 2.94 & 128.7 & 96.8 & 0.042 & 0.968 \\
    w/o Noise in Static Exp. & 2.56 & 118.2 & 96.3 & 0.039 & 0.969 \\
    w/o Relighting Aug. & 2.61 & 106.5 & 98.1 & 0.058 & 0.952 \\
    w/o Visibility Gate & 2.83 & 92.4 & 97.6 & 0.041 & 0.969 \\
    w/o Latent Transport & 5.68 & 138.9 & 94.5 & 0.072 & 0.938 \\
    \midrule
    \textit{Curriculum Variants} & & & & & \\
    Synthetic Only & 2.14 & 154.2 & 98.4 & 0.061 & 0.948 \\
    Real Only & 3.82 & 102.6 & 97.3 & 0.045 & 0.965 \\
    w/o Generative Data & 2.53 & 98.1 & 98.2 & 0.043 & 0.966 \\
    \midrule
    \textbf{Full Model} & \textbf{2.39} & \textbf{79.8} & \textbf{98.8} & \textbf{0.036} & \textbf{0.975} \\
    \bottomrule
  \end{tabular}
\end{table}

\begin{figure}[t!]
\centering
\begin{minipage}[t]{0.49\linewidth}
\centering
\includegraphics[width=\linewidth]{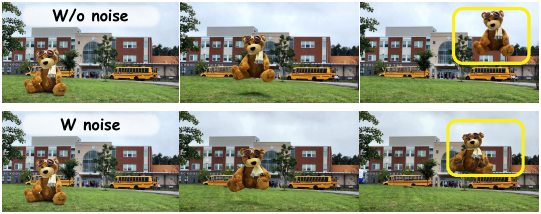}
\caption{\textbf{Visual Ablation Analysis.} Impact of noise injection in static image expansion.}
\label{fig:ablation_noise}
\end{minipage}
\hfill
\begin{minipage}[t]{0.49\linewidth}
\centering
\includegraphics[width=\linewidth]{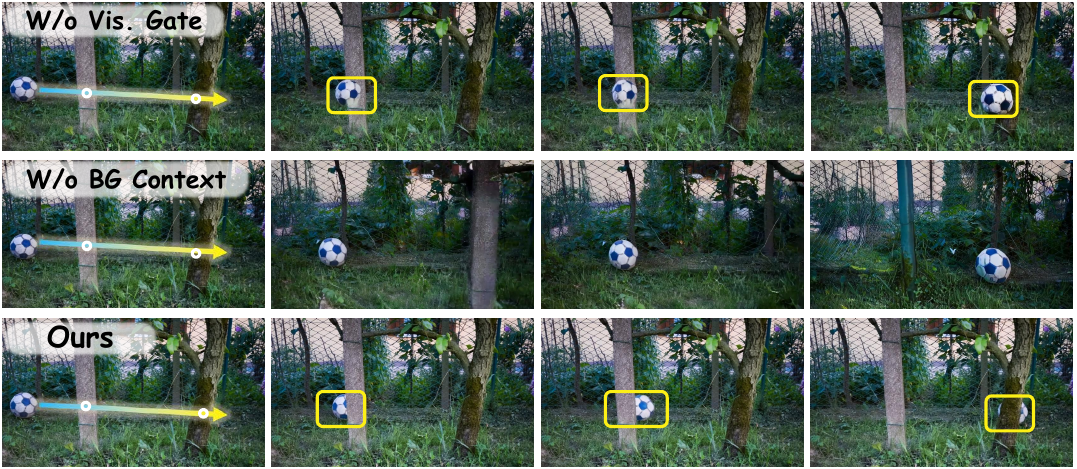}
\caption{\textbf{Visual Ablation on Context and Visibility.} Impact of the Background Context and the visible gate in latent injection module.}
\label{fig:ablation_visible_gate}
\end{minipage}
\end{figure}

\begin{figure*}[t]
    \centering
    \includegraphics[width=1.0\linewidth]{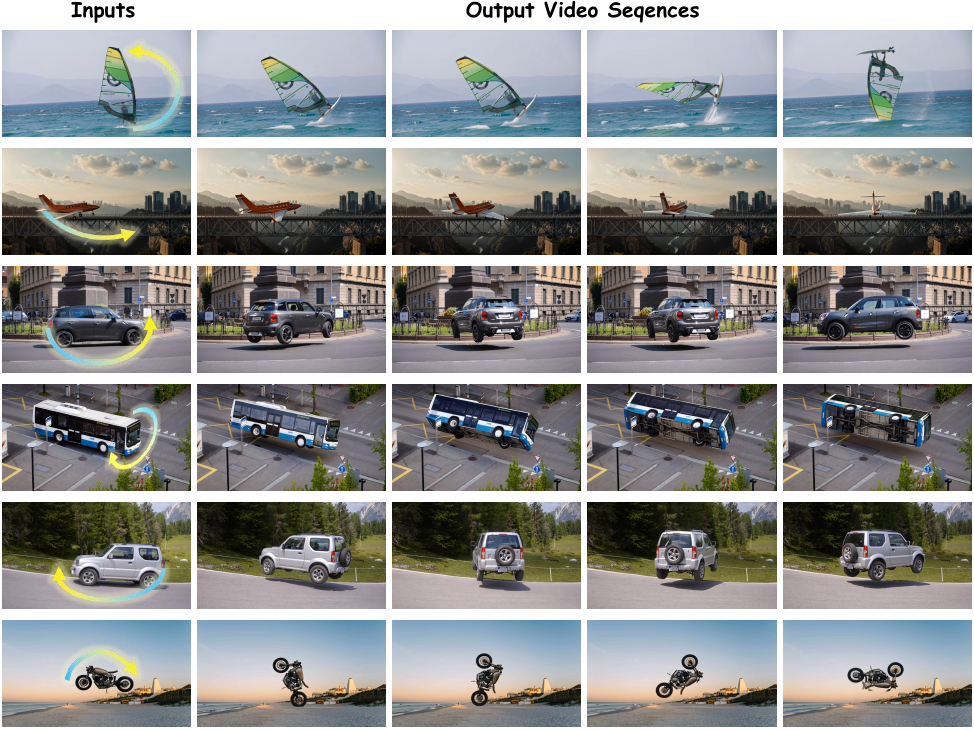}
    \caption{Qualitative results of complex 3D rotations. FlexComposer synthesizes challenging out-of-plane motions (e.g., flips, rolls) by applying explicit 3D transformations to a SAM3D-reconstructed proxy. Injecting the resulting trajectory ensures temporally consistent synthesis even during extreme pose changes.}
\label{fig:supp_3d_rot}
\end{figure*}

\begin{figure*}[ht]
    \centering
    \includegraphics[width=1.0\linewidth]{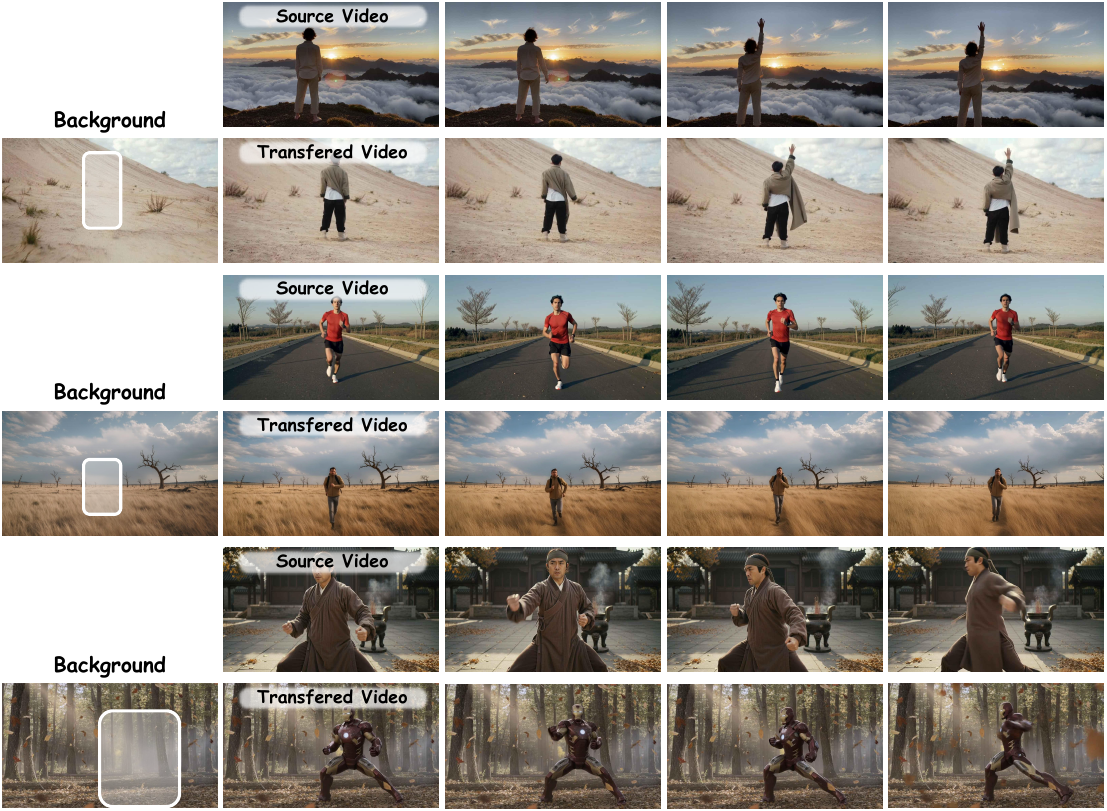}
    \caption{Qualitative results of motion transfer. By mapping source trajectories to target characters via latent transport, our framework facilitates motion synthesis in new background. This approach supports non-pixel-aligned transfer, allowing for motion transfer across different character and backgrounds.}
\label{fig:supp_motion}
\end{figure*}

\subsection{Ablation Study}
To validate the effectiveness of our design choices, we conduct a comprehensive ablation study. We remove key components and analyze their impact on trajectory accuracy (EPE), temporal consistency (FVD), motion preservation (Motion Score from VBench), and harmonization quality (LPIPS, CLIP Score). 
Results are summarized in Table~\ref{tab:ablation}.
More details can be found in supplementary materials.

\noindent\textbf{Canonical Representation \& Noise.} Removing the canonical pipeline (\textit{w/o Canonical Repr.}) causes severe motion conflicts, degrading trajectory adherence. For static inputs, disabling noise injection (\textit{w/o Static Noise}) results in rigid, statue-like outputs.
As shown in Figure~\ref{fig:ablation_noise}, noise is essential for synthesizing plausible micro-motions.

\noindent\textbf{Latent Injection \& Visibility.} 
%
This core module governs the precise placement and integration of the asset into the scene, relying on three components. 
First, the Latent Transport mechanism aligns canonical features with the trajectory. 
Without it (\textit{w/o Latent Transport}), sliding artifacts emerge as objects fail to anchor to the scene geometry.
Second, as shown in Figure~\ref{fig:ablation_visible_gate} (middle), omitting this context (\textit{w/o BG Context}) compromises environmental stability, leading to severe hallucinations in non-edited regions. 
Finally, the Visibility Gate manages the interactions; removing it (\textit{w/o Vis. Gate}) results in ghosting artifacts (Figure~\ref{fig:ablation_visible_gate}, top), where objects incorrectly blend with occluders rather than disappearing behind them.

\noindent\textbf{Relighting Augmentation \& Curriculum.} 
Removing the relighting module leads to a sticker effect with flat lighting. 
While trajectory metrics remain strong, the perceptual quality significantly degrades as objects appear inconsistent with the target scene lighting.
Regarding data, \textit{Synthetic Only} yields precise control but poor realism, while \textit{Real Only} suffers from noisy trajectory signals. 
Our hybrid curriculum effectively balances geometric precision with photorealistic harmonization.

\subsection{Downstream Applications}
FlexComposer's unified representation and flexible trajectory control naturally enable advanced video editing tasks beyond standard compositing.

\noindent\textbf{Complex 3D Motion Synthesis.} 
Our framework can synthesize challenging out-of-plane motions, such as flips, rolls, and axial rotations (see Figure \ref{fig:supp_3d_rot}). We first reconstruct a 3D proxy of the target asset using SAM3D. By applying explicit 3D transformations to this proxy, we extract a precise 2D motion guidance trajectory. Injecting this signal ensures temporally consistent synthesis even during extreme pose changes.

\noindent\textbf{Trajectory-Driven Motion Transfer.} 
By decoupling intrinsic asset features from global spatial coordinates, FlexComposer intrinsically supports non-pixel-aligned motion transfer (see Figure \ref{fig:supp_motion}). We can extract the trajectory from source video and seamlessly map it to an entirely different target subject in a new background. 
This achieves robust motion retargeting across diverse subjects and scenes without requiring rigid alignment or specific fine-tuning.

\section{Conclusions \& Limitations}
We presented \textbf{FlexComposer}, a unified framework that bridges the gap between precise geometric controllability and photorealistic video compositing. By decoupling motion from appearance via our Canonical Representation and Latent Transport, FlexComposer achieves precise trajectory adherence for both static and dynamic assets while maintaining high visual fidelity. Extensive experiments demonstrate that our method effectively balances motion precision with photometric harmonization, establishing a robust foundation for controllable video synthesis. 
Despite these advancements, practical challenges remain to be addressed. 
While lighting is harmonized, physical interactions are not simulated and rely solely on diffusion priors, leading to implausible physics.
Besides, generating long-duration videos with extreme viewpoint changes can suffer from inconsistency.
Future work will explore integrating lightweight physics engines to guide the diffusion process and leveraging 3D representations to enhance consistency and temporal stability.

\label{sec:conclusion}

\bibliographystyle{splncs04}
\bibliography{main}

\clearpage
\appendix

\setcounter{page}{1}
\setcounter{figure}{0}
\setcounter{table}{0}
\renewcommand{\thesection}{\Alph{section}}
\renewcommand{\thefigure}{S\arabic{figure}}
\renewcommand{\thetable}{S\arabic{table}}

\section{Implementation Details}
\label{sec:supp_impl}

In this section, we provide precise specifications regarding the architecture, optimization strategy, and evaluation protocols to facilitate reproducibility.

\subsection{Architectural Specifications}
Our framework leverages the Wan2.1-I2V-14B backbone~\cite{wan2025wan} as the core generative engine. To strictly preserve the pre-trained open-domain priors, we freeze all original parameters of the Variational Autoencoder (VAE) and the Diffusion Transformer (DiT). 
To introduce spatial controllability, we employ Low-Rank Adaptation (LoRA)~\cite{hu2022lora} with a rank $r=64$ and an alpha scaling factor of 32. These learnable matrices are injected into the \textit{Query}, \textit{Key}, \textit{Value}, and \textit{Output} projection layers of every Self-Attention and Cross-Attention block within the DiT. 
This configuration strikes a balance between trainability and expressiveness, allowing the model to learn trajectory-guided feature transport without disrupting the pre-trained motion manifold.

\subsection{Optimization and Parallelism}
We optimize the model using the AdamW optimizer with $\beta_1=0.9, \beta_2=0.999$, a weight decay of $1 \times 10^{-3}$, and a constant learning rate of $1 \times 10^{-5}$. The training is distributed across a cluster of 32 NVIDIA A100 (80GB) GPUs. 
Given the substantial memory footprint of the 14B parameter model and the requirement for long temporal contexts (up to 128 frames), we utilize a hybrid parallelism strategy:
\begin{itemize}
    \item \textbf{Data Parallelism:} Fully Sharded Data Parallel (FSDP) shards model parameters, gradients, and optimizer states across GPUs.
    \item \textbf{Sequence Parallelism:} We adopt Ulysses parallelism with a degree of 4. This splits the spatiotemporal sequence along the attention head dimension, ensuring efficient memory usage and computation for high-resolution video training.
\end{itemize}

\subsection{Dynamic Trajectory Sampling \& Annealing}
\label{sec:supp_sampling}

A critical component of our training strategy—specifically during the \textit{Real-World Adaptation} phase—is balancing precise trajectory adherence with the base model's intrinsic generative priors. We design a dual-mechanism sampling strategy:

\noindent\textbf{Probabilistic Motion Dropout.}
To prevent catastrophic forgetting of the foundation model's Image-to-Video (I2V) capabilities, we employ a probabilistic dropout mechanism. For each iteration, the number of trajectory points $k$ is sampled as:
\begin{equation}
    k \sim 
    \begin{cases} 
    0 & \text{with } p=0.05 \quad (\text{Motion Dropout}) \\
    \hat{k} & \text{with } p=0.95 
    \end{cases}
\end{equation}
When $k=0$, the motion condition is explicitly nulled, compelling the model to rely solely on image contexts to synthesize natural dynamics.

\noindent\textbf{Dense-to-Sparse Annealing Schedule.}
When motion conditioning is active ($k > 0$), we implement a \textbf{dense-to-sparse annealing schedule}. This curriculum progressively adapts the model to varying levels of guidance sparsity. Let $t$ be the current training step and $T_{total}$ be the total steps for Phase 2. The sampling upper bound $K_{max}(t)$ decays from a dense set to a sparse subset:
\begin{equation}
    \hat{k} \sim \mathcal{U}(1, K_{max}(t)),  K_{max}(t) = \left\lfloor 200 \cdot \left(1 - \frac{t}{T_{total}}\right) + 20 \cdot \frac{t}{T_{total}} \right\rfloor
\end{equation}
In early stages, dense trajectories ($\sim$200 points) facilitate rapid learning of the geometric mapping between 2D tracks and latent features. As training progresses, the reduced density ($\sim$20 points) forces the model to interpolate motion between sparse keypoints, ensuring robustness to minimal user inputs during inference.

\subsection{Evaluation Protocol}
To ensure fair comparison with text-driven methods (e.g., Tora), we generate detailed captions from reference images as prompts. For instruction-based editing methods (e.g., AnyV2V), we construct specific editing instructions aligned with the ground-truth intent.
\begin{itemize}
    \item \textbf{Quantitative Metrics:} \textit{End-Point Error (EPE)} calculates the average Euclidean distance between the tracked centroid (via SpatialTrackerV2) and the ground truth trajectory, normalized by the image diagonal.
    \item \textbf{Perceptual Metrics:} We adopt the VBench protocol to report Subject Consistency, Motion Smoothness, and Aesthetic Quality.
\end{itemize}

\begin{figure}[t!]
    \centering
    \includegraphics[width=1.0\linewidth]{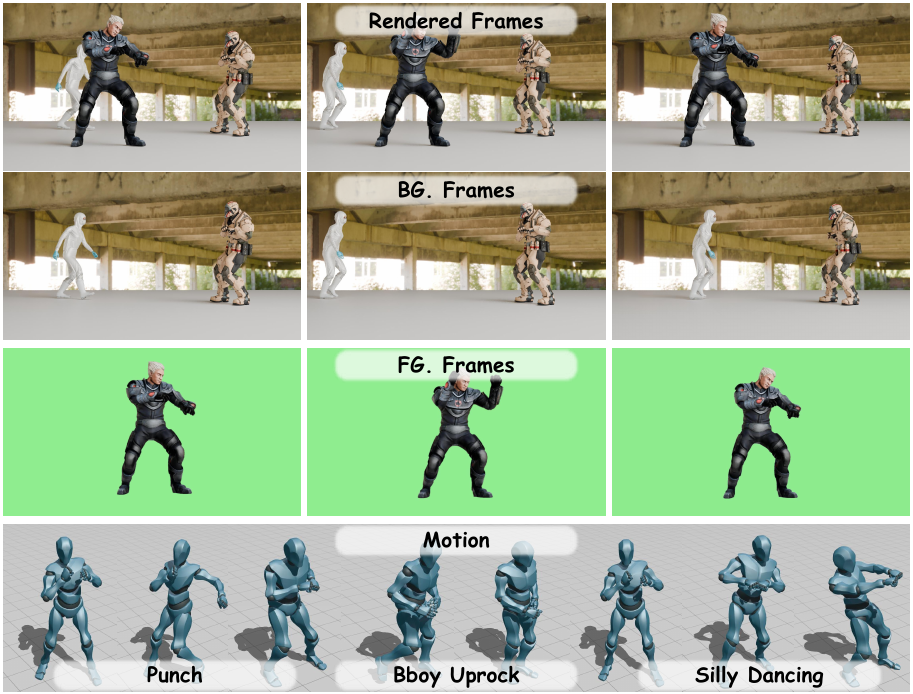}
    \vspace{-2mm}
    \caption{Synthetic Data Generation and Multi-Character Protocol. This figure illustrates our data synthesis pipeline designed to disentangle motion dynamics from visual appearance. (Top Row) Rendered Frames: The final high-fidelity composite featuring realistic lighting and multiple characters. (Second Row) BG. Frames: Dynamic background environments that include other moving characters as "distractors." (Third Row) FG. Frames: The isolated target foreground character. (Bottom Row) Motion: Representative motion sequences from the Mixamo library. By designating only one target subject while treating others as dynamic background elements, this protocol forces the model to learn robust motion preservation and discrimination in complex, cluttered environments.}
    \vspace{-2mm}
\label{fig:supp_3d_rot}
\end{figure}

\begin{figure*}[t!]
    \centering
    \includegraphics[width=0.7\linewidth]{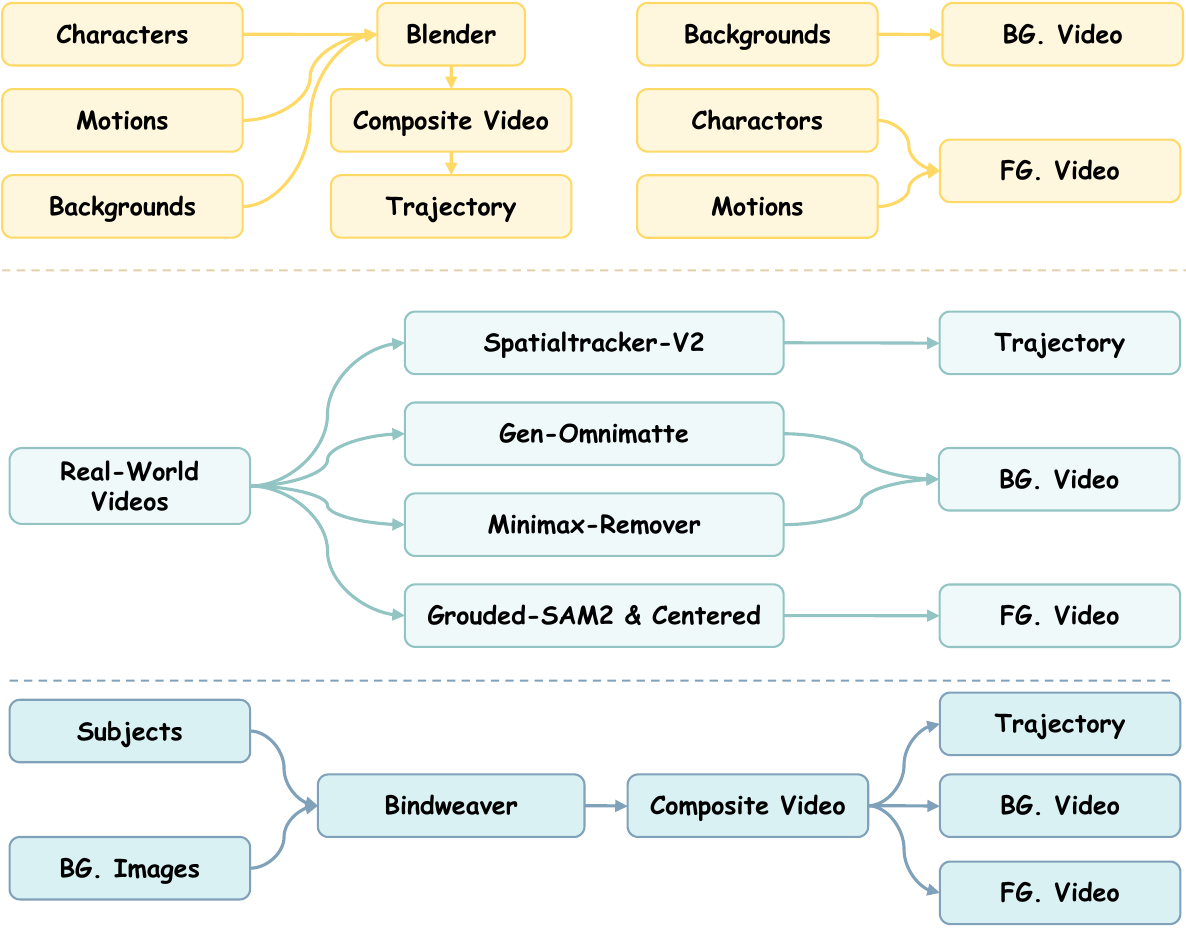}
    \caption{Illustration of our data pipeline.}
\label{fig:supp_data_pipe}
\end{figure*}

\begin{figure*}[ht]
    \centering
    \includegraphics[width=0.90\linewidth]{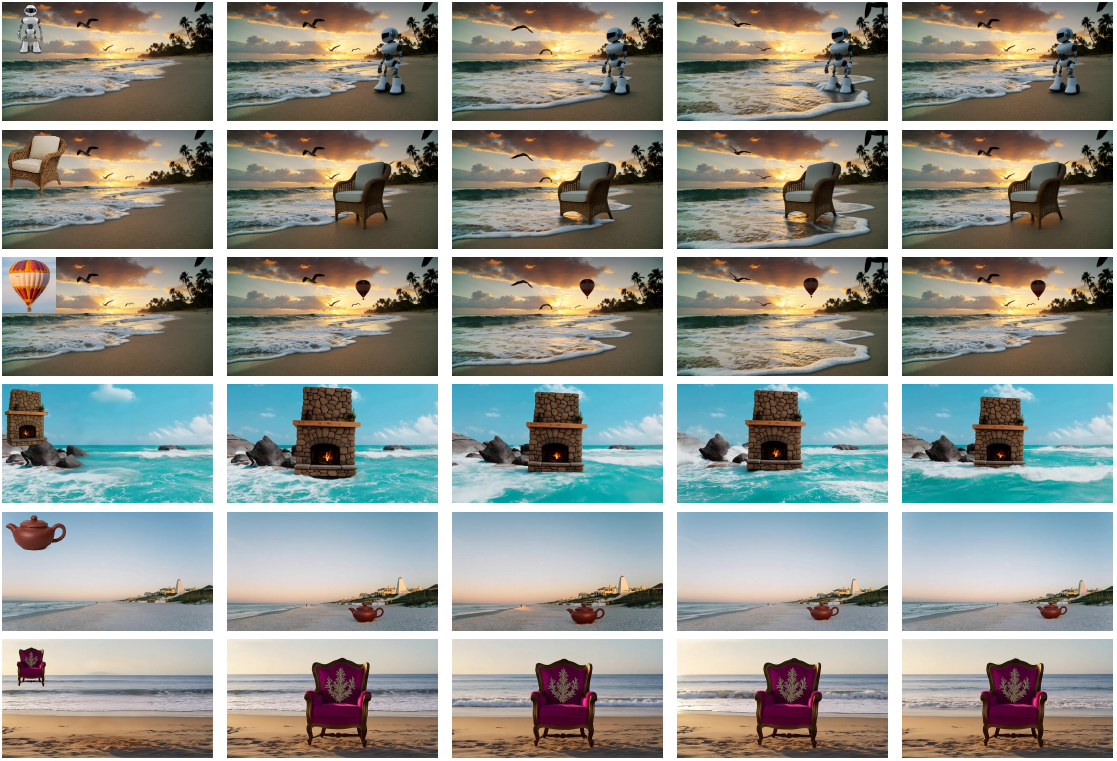}
    \vspace{-2mm}
    \caption{Additional qualitative results of video harmonization.}
    \vspace{-2mm}
\label{fig:supp_3d_rot}
\end{figure*}

\section{Dataset Curation and Processing}
\label{sec:supp_dataset}

Our three-stage curriculum relies on a diverse mixture of synthetic, real-world, and generative data, processed through a rigorous pipeline to ensure geometric and semantic quality.

\subsection{Phase 1: Synthetic Data with Dynamic Environments}
To facilitate the \textit{Simulation Bootstrap} phase, we engineered a procedural dataset aimed at disentangling motion dynamics from visual appearance.

\noindent\textbf{Environment \& Lighting.} We synthesize high-fidelity backgrounds using Kubric~\cite{greff2022kubric} assets combined with HDRI maps from Polyhaven. This ensures physically plausible shadows and reflections, narrowing the domain gap with real-world illumination.

\noindent\textbf{Multi-Character Protocol.} We utilize the Mixamo library~\cite{mixamo} (40 characters, 582 sequences). In each generation, 1 to 5 characters are instantiated with independent trajectories. Crucially, only one character is designated as the target subject, while others act as \textit{dynamic distractors}. This forces the network to learn unsupervised motion preservation and distinguish the target's dynamics from environmental clutter.

\subsection{Phase 2: Real-World Adaptation Pipeline}
We construct the primary training set from diverse Pexels footage. To transform raw videos into high-quality training pairs, we implement a four-stage pipeline:

\noindent\textbf{1. Filtration.} We employ \textbf{PySceneDetect} to enforce temporal continuity, partitioning sequences into 81--100 frame clips. We then apply a feature-based displacement metric (ORB features and RANSAC) to eliminate static or low-information scenes, and further filter out low-luminance clips and facial close-ups.

\noindent\textbf{2. Geometric Extraction.} We employ SpatialTrackerV2~\cite{xiao2024spatialtracker} to lift 2D pixel dynamics into 3D space, providing dense trajectories and 6DoF camera poses. Simultaneously, we use the Grounded-SAM2 pipeline~\cite{Ren2024GroundedSA}—combining Grounding DINO and SAM—to generate pixel-perfect binary masks for the primary moving subjects.

\noindent\textbf{3. Background Reconstruction.} Using the masks and camera tracks, we hallucinate clean background plates. We prioritize scale via a hybrid strategy: Gen-Omnimatte~\cite{lee2025generative} is used for complex occlusion decomposition, while the computationally efficient Minimax-Remover~\cite{Zi2025MiniMaxRemoverTB} handles the majority of inpainting. We deliberately retain samples with minor inpainting artifacts to serve as data regularization, enhancing robustness against occlusions.

\noindent\textbf{4. Semantic Annotation.} We employ \textbf{Keye-VL} to generate dense semantic annotations describing scene layout and temporal evolution. These motion-aware captions enable the cross-attention mechanism to align textual verbs with induced motion trajectories.

\subsection{Phase 3: Generative Diversity Expansion}
To support open-domain generalization, we synthesize $\sim$5,000 clips using state-of-the-art T2V models based on imaginative prompts (e.g., ``cyberpunk animals'', ``underwater physics''). Pseudo-ground-truth trajectories are estimated via the same tracking pipeline as Phase 2. This strategy prevents overfitting to specific object categories found in real-world footage.

\section{Detailed Ablation Settings and Analysis}
\label{sec:extended_ablation}

To provide a comprehensive understanding of the design choices in FlexComposer, we detail the exact configurations for the ablation variants reported in the main paper and analyze the implications of removing these components.

\subsection{Canonical Representation and Preprocessing}

\noindent\textbf{w/o Canonical Representation.} 
This variant disables the \textit{Dynamic Video Stabilization} step. Raw, cropped foreground video clips are fed directly into the VAE. Without our stabilization-then-transport logic, the model fails to reconcile the conflicting ego-motion of the source video with the target trajectory, leading to severe motion conflicts and sliding artifacts.

\noindent\textbf{Static Image Expansion Strategies.}
To unify static and dynamic inputs, our full model expands images into pseudo-sequences with Gaussian noise injection. We analyze two degraded settings:
\begin{itemize}
    \item \textbf{w/o Static Expansion:} Treats the static image as a single-frame condition broadcasted implicitly.
    \item \textbf{w/o Noise Injection:} Replicates the static image $T$ times without noise ($I_{expand}[t] = I_{raw}$). 
\end{itemize}
Simply replicating frames often causes the backbone to produce a frozen video ("statue effect"). Our noise injection strategy breaks this redundancy, encouraging the model to hallucinate plausible micro-motions (e.g., breathing, fabric movement) while adhering to the global trajectory.

\noindent\textbf{w/o Relighting Augmentation.}
This configuration disables random illumination perturbations and color jittering. Without this module, the model defaults to a "copy-paste" behavior, retaining the original lighting of the source asset (the "sticker effect"). Our augmentation effectively forces the model to predict lighting parameters implicitly from the background context to minimize reconstruction loss.

\subsection{Spatial-Aware Injection Mechanism}

\noindent\textbf{w/o Latent Transport (Learned Adapter).} 
We replace our parameter-free injection with a learnable convolutional encoder that processes trajectory heatmaps and foreground features. Results indicate that while a learnable encoder can approximate motion, it suffers from signal degradation and "sliding" artifacts. Our explicit geometric transport mathematically enforces alignment without auxiliary parameters, ensuring superior trajectory adherence.

\noindent\textbf{w/o Visibility Gate.}
Setting the visibility term $\mathcal{V} \equiv 1$ globally ignores depth cues. This leads to ghosting artifacts where objects incorrectly blend with occluders rather than disappearing behind them.

\noindent\textbf{w/o BG Context.}
Removing the background latent $Z_{bg}$ from the fusion step tasks the model with generating the background from scratch. This compromises environmental stability, leading to hallucinations in non-edited regions.

\subsection{Curriculum Learning Strategy}

We analyze the impact of our curriculum stages by isolating specific data sources.
\textbf{Synthetic Only} trains exclusively on procedural simulation clips. While it benefits from perfect ground-truth supervision and achieves high geometric precision (low EPE), the model suffers from a severe \textit{sim-to-real domain gap}. 
Consequently, it fails to synthesize photorealistic textures or complex lighting interactions when applied to real-world footage, often yielding composites with artificial, "game-like" appearances. 
Conversely, the \textbf{Real Only} variant bypasses the synthetic warm-up stage. Without the strong initial geometric priors established by clean synthetic data, the model is forced to learn motion control directly from estimated trajectories, which inherently contain noise and occlusion errors. This "noisy supervision" prevents the model from effectively disentangling intrinsic object motion from global displacement, resulting in jittery trajectories and reduced control fidelity. 
Finally, \textbf{w/o Generative Data} omits the open-domain refinement phase. While the model maintains robust performance on common categories present in the real-world dataset (e.g., humans, vehicles), it exhibits poor generalization to out-of-distribution or imaginative scenarios. 
This highlights the critical role of diverse generative data in expanding the model's semantic coverage and preventing overfitting to limited real-world distributions.

\begin{table}[t]
\centering
\caption{\textbf{User Study Statistics.} Preference rate (\%) of Ours vs. Baselines. Our method demonstrates superior control and visual integration, maintaining competitive performance even against state-of-the-art motion models.}
\label{tab:supp_user_study}
\begin{tabular}{l|cccc}
\toprule
\textbf{Comparison Pair} & \textbf{Trajectory} & \textbf{Harmony} & \textbf{Fidelity} & \textbf{Overall} \\
\midrule
\multicolumn{5}{l}{\textit{Task 1: Static Foreground Compositing}} \\
Ours vs. Tora & 82.4\% & 77.6\% & 65.2\% & 75.1\% \\
Ours vs. Wan-Move & 53.2\% & 55.8\% & 51.4\% & 53.5\% \\
Ours vs. Kling 1.5 & 87.6\% & 57.2\% & 55.4\% & 66.7\% \\
\midrule
\multicolumn{5}{l}{\textit{Task 2: Dynamic Foreground Compositing}} \\
Ours vs. AnyV2V & 79.2\% & 84.4\% & 87.6\% & 83.7\% \\
Ours vs. GenCompositor & 71.6\% & 66.4\% & 75.2\% & \textbf{71.1\%} \\
\bottomrule
\end{tabular}
\end{table}

\section{Additional User Study Results}
\label{sec:supp_user_study}
To  assess perceptual quality and controllability, we conducted a Two-Alternative Forced Choice (2AFC) user study. 
The study panel consisted of 25 independent evaluators. 
We utilized a diverse dataset of 20 randomly selected test cases covering a spectrum of motion complexities, ranging from simple 2D translations to challenging 3D rotations. 
In each trial, participants were presented with the reference inputs (Trajectory, Asset, and Background) alongside two anonymized video results (Ours vs. Baseline) displayed side-by-side in randomized order. 
Evaluators were asked to select the preferred video based on three specific criteria: Trajectory Adherence ("Which video better follows the indicated arrow path?"), Visual Harmony ("Which video shows more realistic lighting, shadows, and blending?"), and Identity Fidelity ("Which video better preserves the details of the reference object?").

As summarized in Table~\ref{tab:supp_user_study}, our method achieves comparable performance with Wan-Move, the leading motion-controlled model, while showing a clear user preference for \textbf{Visual Harmony} (55.8\%) due to our canonical representation's handling of lighting. 
Against commercial models like Kling 1.5, which prioritize per-frame quality over control, we maintain a dominant lead in \textbf{Trajectory Adherence} (87.6\%). In dynamic compositing tasks, FlexComposer significantly outperforms AnyV2V and GenCompositor, particularly in \textbf{Identity Fidelity} (87.6\% vs. AnyV2V), by avoiding the texture flickering and distortion common in video-to-video translation pipelines.

\section{Failure Case Analysis}
\label{sec:supp_failure_case}

\begin{figure}[t!]
    \centering
    \includegraphics[width=1.0\linewidth]{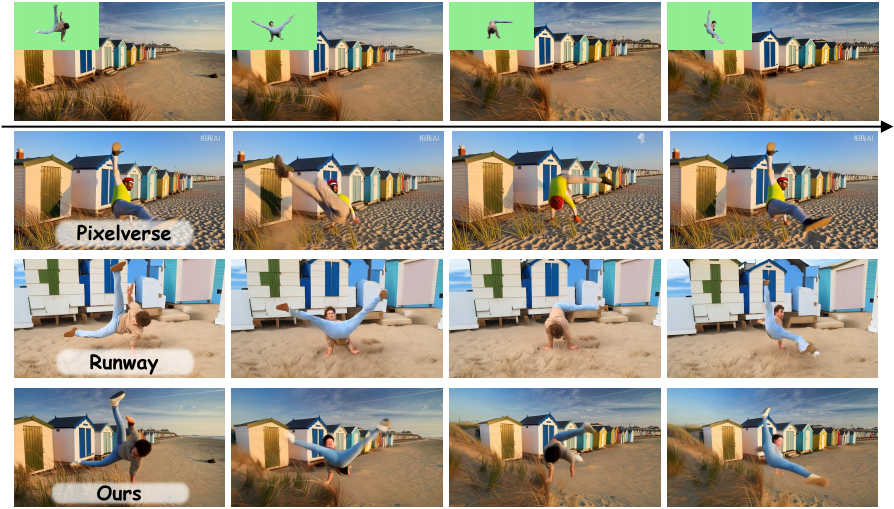}
    \caption{ \textbf{Failure case analysis on large motion.} We compare methods on a challenging breakdancing sequence. Commercial models struggle with consistency: \textbf{Pixelverse} exhibits identity drift, while \textbf{Runway} alters the background layout. 
  In contrast, our method preserves better structural and identity fidelity, though it suffers from motion blur on fast-moving limbs due to the backbone's limitation in handling rapid dynamics.}
\label{fig:supp_failure_case}
\end{figure}


We investigate the robustness of our model on scenarios with rapid, large-scale body motions (e.g., breakdancing). As illustrated in Fig.~\ref{fig:supp_failure_case}, such extreme dynamics present a challenge for both our method and commercial tools.
Commercial video generation models (e.g., Pixelverse, Runway) tend to prioritize motion smoothness at the cost of consistency. Pixelverse often exhibits identity drift, while Runway frequently hallucinates background details deviating from the input layout.
In comparison, our method maintains higher fidelity regarding background structure and subject identity due to explicit compositional constraints. However, it struggles with high-frequency details during rapid articulation, occasionally resulting in motion blur or geometric artifacts on limbs. This suggests that while our approach effectively anchors global structure, the underlying priors face difficulties resolving sharp details under extreme motion magnitude.

\section{Discussion of Concurrent Works}
\label{sec:supp_concurrent}
The field of generative video editing is evolving rapidly. 
Several works have been developed concurrently with FlexComposer.
Recently, several approaches have explored trajectory-based video manipulation. 
MotionV2V and Edit-by-Track focus on editing existing scene dynamics (e.g., altering the path of a car already present in the video). 
In contrast, FlexComposer addresses the challenge of compositing, which necessitates harmonizing external assets into a new environment. Unlike editing approaches that leverage existing pixel context, our framework explicitly learns to synthesize plausible lighting and shadows for novel objects via our hybrid synthetic-to-real curriculum.
Closer to our goal, GenCompositor and WorldCanvas explore generative object insertion. 
However, GenCompositor primarily relies on learned adapters that often struggle to preserve the intrinsic motion fidelity of complex pre-animated assets (as analyzed in Sec. 4.3). 
Similarly, while WorldCanvas offers a powerful general-purpose simulator, our method is specifically tailored for precise compositing workflows. 
Our key distinction lies in the Unified Canonical Foreground Representation and Parameter-free Latent Injection, which effectively decouple intrinsic object dynamics from global trajectory control. 
This design allows FlexComposer to transport dynamic footage with high fidelity—preserving micro-motions and identity—while enforcing strict geometric adherence, a capability that remains challenging for generalist world models.

\end{document}